\documentclass[10pt,twocolumn,letterpaper]{article}

\usepackage{cvpr}              
\definecolor{cvprblue}{rgb}{0.21,0.49,0.74}
\usepackage[pagebackref,breaklinks,colorlinks,allcolors=cvprblue]{hyperref}
\usepackage{tabu}                     
\usepackage{multirow}                
\usepackage{float}                    
\usepackage{makecell}                 
\usepackage{booktabs}                 
\usepackage{colortbl}
\usepackage{setspace}
\usepackage{bm}
\usepackage{xcolor}

\def\paperID{2361} 
\def\confName{CVPR}
\def\confYear{2025}

\title{Re-HOLD: Video Hand Object Interaction Reenactment via adaptive Layout-instructed Diffusion Model}

\author{
    Yingying Fan$^1$$^{\dagger}$ \and 
    Quanwei Yang$^2$ \and
    Kaisiyuan Wang$^3$$^{\ast}$ \and 
   Hang Zhou$^3$ \and 
   Yingying Li$^3$ \and 
   Haocheng Feng$^3$ \and 
    Errui Ding$^3$ \and
   Yu Wu$^{1}$$^{\ast}$\and 
   Jingdong Wang$^3$ \and
     $^{1}$ School of Computer Science, Wuhan University \qquad \\
     $^{2}$ University of Science and Technology of China \qquad
     $^{3}$ Baidu Inc.\\
}
\begin{document}

\twocolumn[{
\renewcommand\twocolumn[1][]{#1}%
\maketitle
\begin{center}
 \centering
 \includegraphics[width=\textwidth]{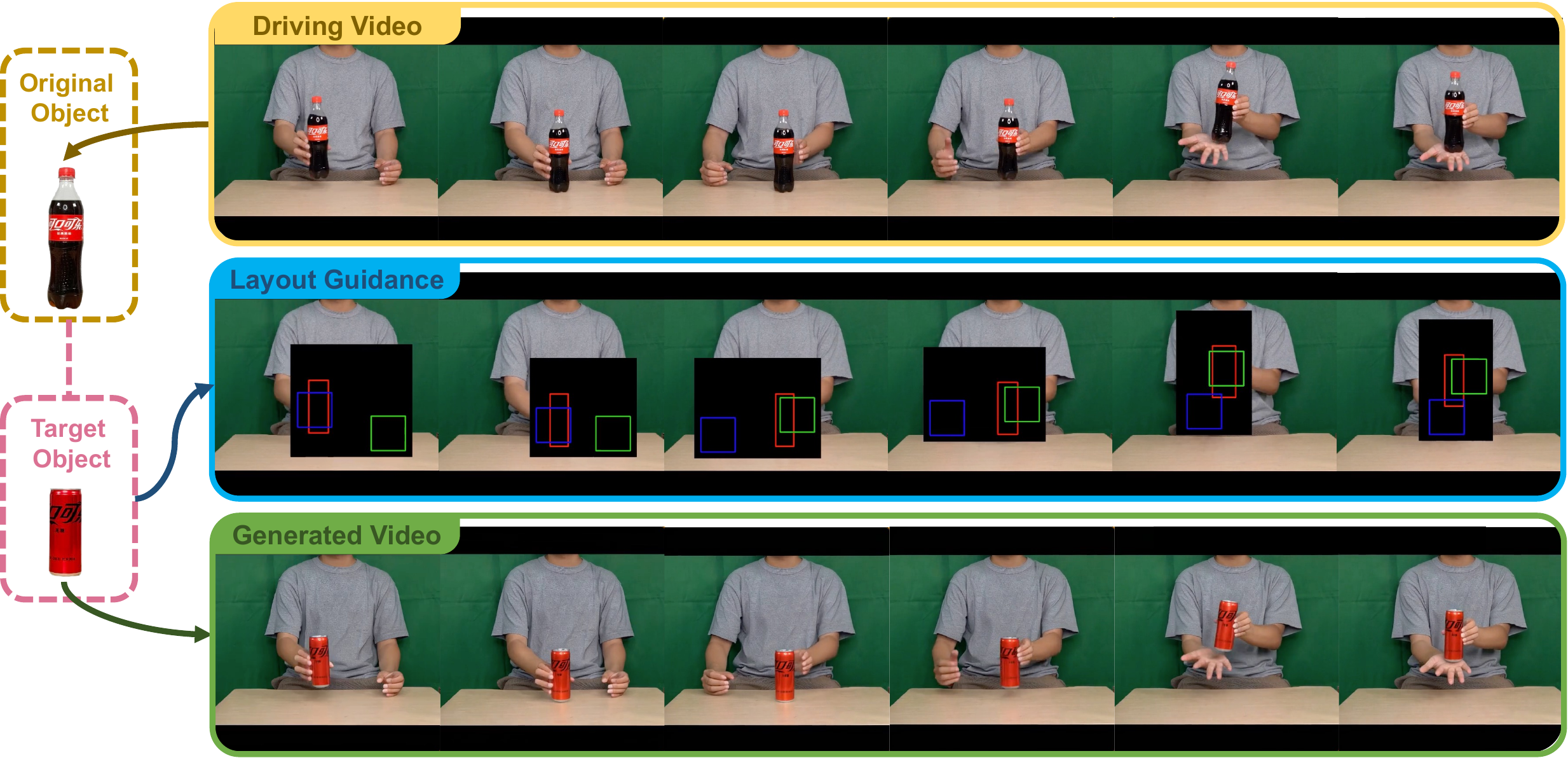}
\captionof{figure}{
\textbf{Cross-Reenactment Results by our Re-HOLD framework.} Given a driving video and a target object, Re-HOLD can synthesize high-fidelity Human-Object Interaction (HOI) videos, even when the sizes of the target and original object differ significantly. 
}
\label{fig:teasor}
\end{center}
}]

\renewcommand{\thefootnote}{\fnsymbol{footnote}}
\footnotetext[2]{Work done during an internship at Baidu Inc.}
\footnotetext[1]{Corresponding author}

\begin{abstract}
Current digital human studies focusing on lip-syncing and body movement are no longer sufficient to meet the growing industrial demand, while human video generation techniques that support interacting with real-world environments (e.g., objects) have not been well investigated.
Despite human hand synthesis already being an intricate problem, generating objects in contact with hands and their interactions presents an even more challenging task, especially when the objects exhibit obvious variations in size and shape. 
To tackle these issues, we present a novel video reenactment framework focusing on Human-Object Interaction (HOI) via an adaptive Layout-instructed Diffusion model (Re-HOLD).
Our key insight is to employ specialized layout representation for hands and objects, respectively. 
Such representations enable effective disentanglement of hand modeling and object adaptation to diverse motion sequences.
To further improve the quality of the HOI generation, we design an interactive textural enhancement module for both hands and objects by introducing two independent memory banks.
We also propose a layout adjustment strategy for the cross-object reenactment scenario to adaptively adjust unreasonable layouts caused by diverse object sizes during inference.
Comprehensive qualitative and quantitative evaluations demonstrate that our proposed framework significantly outperforms existing methods. Project page: \url{https://fyycs.github.io/Re-HOLD}.
\end{abstract}

\section{Introduction}
\label{sec:intro}

With the rapid advancements in human video generation technology, digital human services have extended their reach into our daily routines (e.g.,  education, e-commerce, and multi-modal entertainment), leading to a notable rise in the demands of digital human videos. In response to these demands, numerous studies~\cite{wav2lip,lsp,vprq,liveportrait,pcavs,LIA,facevid2vid,vpgc,emo,hallo} committed to 2D speaking animation have been proposed to create an interactive conversational experience.

On the other hand, human videos solely limited to lip movements or head movements synthesis fail to deliver a satisfactory user experience in real-world scenarios, which prompts the exploration of video generation for human body movements~\cite{dreampose,tps,animateanyone,magicanimate,magicpose,disco,vpu}. However, these approaches typically concentrate on holistic body motion modeling based on either 2D body poses~\cite{dwpose} or implicit motion representation and take inadequate consideration of human hands, which are essential components in interactive scenarios. Although latest studies~\cite{makeanchor, realisdance,talkact} continue to optimize hand synthesis by involving 3D hand mesh, they still cannot produce compelling results in more interactive scenarios with complex hand-object interaction (HOI).
Nevertheless, HOI synthesis is a highly challenging research area, where even generating HOI images poses considerable difficulties, leading to video-level investigation particularly sparse. 
Thus how to achieve realistic HOI video synthesis remains an open problem.
Previous image-level works~\cite{affordance, graspdiffusion} built upon ControlNet~\cite{controlnet} attempt to generate either an articulated hand or a full-body hand-grasping pose image for a given object.
More recently, HOI-Swap~\cite{hoi-swap} extends image-level HOI inpainting into a video-level framework by leveraging an additional stage for sequential frame warping. However, it can only produce object-centric videos with single-hand grasping operations and limited hand involvement.

Our primary focus is to devise a human-centric video generation system that enables reasonable HOI synthesis according to a source motion sequence and a target object.
However, building such a system is non-trivial, since it entails three challenging problems:
\textbf{1)} The physical interaction between hands and objects usually creates diverse occlusions, which leads to their intricate entanglement and easily causes artifacts at the hand-object interface.
\textbf{2)} Both hands and objects exhibit high degrees of freedom and occupy only limited pixels in each video frame. Solely recovering either of them with detailed textures confronts a significant challenge.
\textbf{3)} The non-negligible differences between distinct objects in shape and size inevitably affect the interacting position and degrade the realism of HOI, if the source motion sequence remains unchanged.

To tackle these problems, we propose a Video \textbf{Re}enactment framework for \textbf{H}and-\textbf{O}bject Interaction via \textbf{L}ayout-instructed \textbf{D}iffusion Model, namely \textbf{Re-HOLD}. Our key insight is to pursue hand-object disentanglement by involving specialized layout representations for hands and objects, respectively. Particularly, the layout representation for a video frame is composed of three detected bounding boxes, where two of them from hands (i.e., blue and green ones in Fig.~\ref{fig:teasor}) are shaped in a fixed size, while the rest (i.e., the red one in Fig.~\ref{fig:teasor}) has a varying size according to the object and depth. Notably, the hand layout representation exhibits \textit{pose-invariance} and \textit{size-invariance} providing merely positional information, which benefits the disentanglement between hands and objects.
While such sparse layout representations may not achieve correct and fine-grained HOI synthesis, the disentanglement they provided allows us to introduce more representative instructions for better generation.
Therefore, we further present a Hand-Object Interaction Restoration module to perform structure reshaping and texture refinement, where the former incorporates 3D hand meshes for better structural guidance and the latter relies on two independent memory banks and corresponding masks.
Considering the gap between diverse objects under the cross-reenactment setting, we novelly design an adaptive strategy for layout adjustment at the inference stage, aiming to avoid producing unreasonable physical contact or interactive position.
Extensive experiments demonstrate that our framework reenacts HOI videos with better fidelity than previous methods.

Our contributions are summarized as follows:
\textbf{1)} We propose the first HOI reenactment framework for human-centric video generation which achieves realistic and reasonable HOI synthesis.
\textbf{2)} Our proposed specialized layout representations for hands and objects along with our HOI Restoration Module, enable effective disentanglement and improved HOI modeling.
\textbf{3)} Our proposed layout adjusting strategy is compatible with diverse objects, even ones with obvious gaps in shape and size, to generate reasonable interactions.
\section{Related Work}

\begin{figure*}[t]
	\centering
\includegraphics[width=1\textwidth]{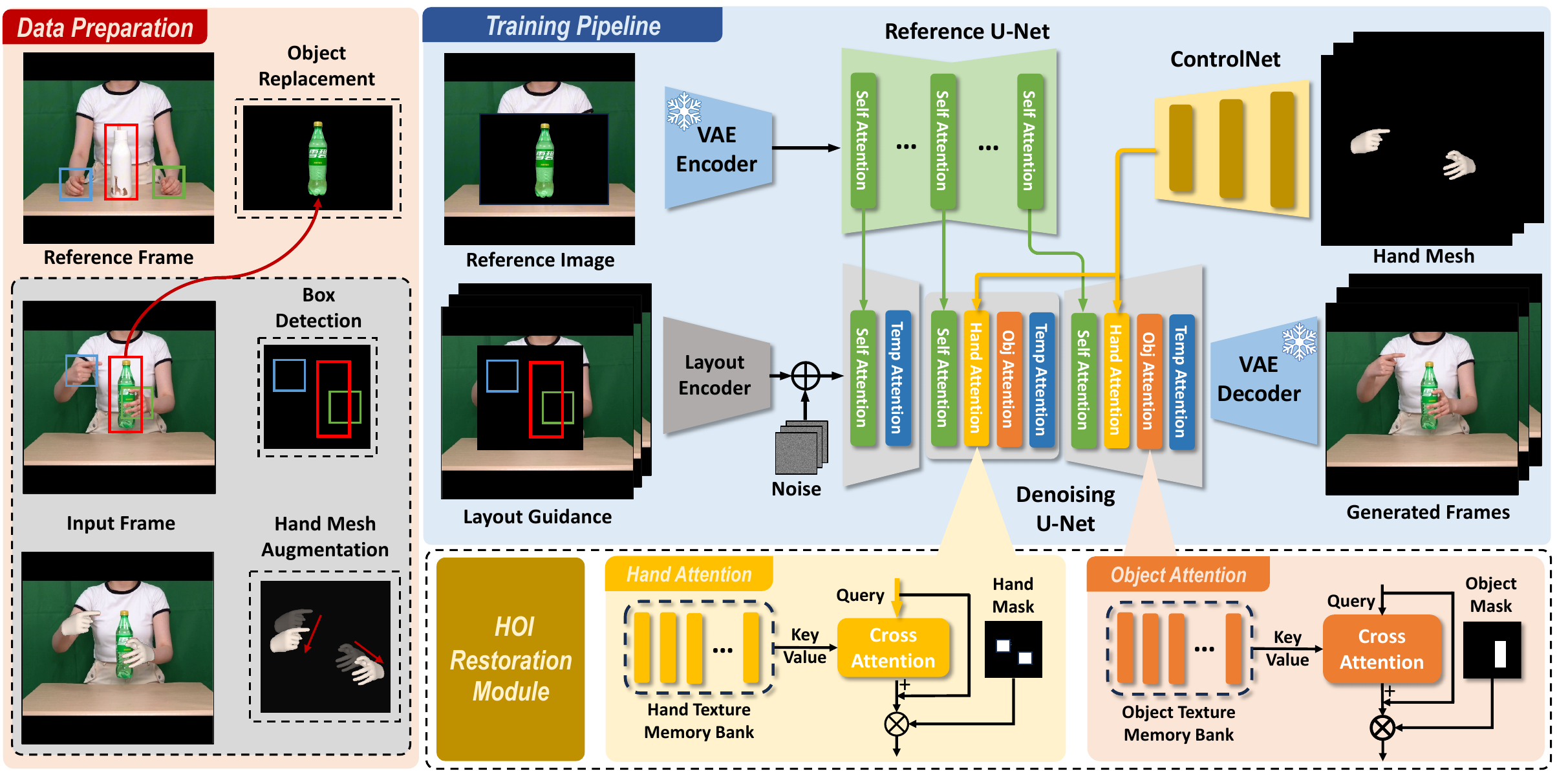} 
	\caption{\textbf{Overview of our proposed Re-HOLD framework}. We propose a two-branch framework that consists of a Reference U-Net and a Denoising U-Net. The Reference U-Net takes a reference object image for object texture encoding while the denoising one takes noise latent and layout guidance as input for diffusion processing. To enhance the quality of HOI generation, we adopt the HOI Restoration Module for hand information and fine-grained object information restoration. }
	\label{pipeline}
\end{figure*}

\subsection{Human Body Animation}
Recent studies have leveraged powerful diffusion-based models to address the challenge of human body animation. One prominent approach involves using a UNet-based network \cite{unet} enhanced with cross-attention mechanisms \cite{attention} to inject additional information. PIDM \cite{pidm} was the first to introduce classifier-free diffusion guidance for pose-guided human image generation, as further developed by \cite{liao}. DreamPose \cite{dreampose} utilizes UV maps as motion signals and performs conditional embeddings to achieve motion transfer. Similar concepts have been explored in other works \cite{animateanyone, disco, magicanimate, champ, magicpose, talkact, yang2025showmaker, icassp}. RealisDance \cite{realisdance} takes DWPose \cite{dwpose}, SMPL \cite{smplx}, and HaMeR \cite{hamer} as input to generate realistic hand poses. While these approaches have yielded promising results, there are challenges when dealing with accurate HOI generations. Recently, InteractDiffusion \cite{interactdiffusion} introduces the layout input to capture intricate interaction relationships. Inspired by this, we employ the sparse layout guidance as the driving signal and adaptively modify the hand-object interaction information during inference.

\subsection{Video Generation and Editing}
Numerous studies have focused on adapting pre-trained diffusion models for video generation and editing, including zero-shot \cite{tokenflow, rerender, pix2video, fatezero, space, vidtome, wang2023diffusion} and one-shot-tuned \cite{tuneavideo, cut, shape, videoswap} learning frameworks. However, these approaches typically demand extensive per-video fine-tuning, which significantly hampers their practical application. Alternatively, other research \cite{structure, control, flowvid, videocomposer, moonshot, vase, omnimatte, zhu2022discrete, liu2024iterative} takes a training-based approach, where models are trained on large datasets to ensure they can serve as efficient editing tools during inference. However, none of these methods specifically address the adaptive generation of human-object interaction, and accurate HOI synthesis has always been a significant challenge in video generation tasks. In addition, a significant proportion of video generation and editing methodologies \cite{stablediffusion, tokenflow, emu, rerender, fatezero, pix2video, space, tuneavideo, shape, ccedit, editavideo, videop2p, makeavideo, videogeneration} primarily depend on textual input for guidance during the editing process. However, textual prompts can sometimes fall short in accurately conveying the user's intentions. For our specific application, utilizing an object image as a form of guidance proves to be more precise and effective.

\subsection{Human-Object Interaction}
Given that human motion rarely occurs in isolation but rather within the context of objects or the surrounding environment, numerous methods \cite{place, synthesizing, populating, generating, resolving} have been developed to explore the realistic integration of humans into scenes. A variety of methods have also focused on more fine-grained hand-object interactions \cite{hoi, affordance, diffusionhoi, hoidiffusion, cghoi, graspxl, lego}. DiffHOI \cite{diffusionhoi} proposes a diffusion network to model the conditional distribution of geometric renderings of objects and leverage it to guide the novel-view rendering. GraspXL \cite{graspxl} unifies the generation of hand-object grasping motions across multiple motion objectives, diverse object shapes, and dexterous hand morphologies. Cg-hoi \cite{cghoi} focuses on generating realistic 3D human-object interactions from textual descriptions. Recently, HOI-Swap \cite{hoi-swap} presents a diffusion-based video editing framework for video object swapping with HOI awareness, ignoring the size and position change of hands and objects during object swapping.
For practical digital human applications, we aim to generate two-hand human-object interactions using an adaptive layout-instructed diffusion model.
\section{Method}
\label{method}

In this section, we first describe our task formulation in Sec.\ref{3.1}. Then we introduce the pipeline of our framework and its important components in Sec.\ref{network}. The training strategy is described in Sec.\ref{train}. The overview of our proposed method is shown in Fig.\ref{pipeline}.

\subsection{Task Formulation}
\label{3.1}
\noindent\textbf{Task Formulation.} HOI reenactment aims to generate reasonable interaction between hands and objects given a sequence of human motion signals and a target object $I_o\in\mathbb{R}^{H \times W \times 3}$. Here, sequential human motion signals in our Re-HOLD framework include layout guidances $V^{l}=\left\{I^{l}_1, I^{l}_2, \ldots, I^{l}_F\right\}\in\mathbb{R}^{F \times H \times W \times 3}$ and reconstructed hand meshes $V^{h}=\left\{I^{h}_1, I^{h}_2, \ldots, I^{h}_F\right\} \in \mathbb{R}^{F \times H \times W \times 3}$.
The training for our framework is performed via a self-reconstruction manner, where the human motion signals and target objects are both from the source video $V=\left\{I_1, I_2, \ldots, I_F\right\} \in \mathbb{R}^{F \times H \times W \times 3}$. Our goal is to reconstruct $V$ from random noise and these two conditional inputs.

At the inference stage, a reference image from another object $I_{o^{'}}$ is provided to reenact the target video $V^{'}$.
To guarantee the realism of the reenactment results, modification of the hand box and object box within the layout is conducted according to the difference between the two objects. Hand poses are kept unchanged to ensure that hands can effectively and adaptively interact with the new object.

\subsection{Re-HOLD Framework Designs}
\label{network}

\noindent\textbf{Baseline Architecture.} Correctly synthesizing hands, particularly fingers, is acknowledged as a quite challenging problem, which becomes even more difficult when it comes to generating in-contact objects and their interactions.
Inspired by recent studies on human animation ~\cite{animateanyone, realisdance, magicpose, talkact}, we initially formulated a baseline built upon a parallel-branch architecture to represent the human motion and the target object separately. 

The upstream branch processes the reference image $I_o$ of the target object to extract texture information by using a VAE encoder and a Reference U-Net.
In the downstream branch, a Motion Encoder takes the human motion signals as input to integrate information about motion and structures. The subsequent Denoising U-Net effectively combines the extracted object texture and motion to predict noise intensity. Additional temporal attention layers are employed to improve temporal coherence.
These two branches interact with each other through cross-attention mechanisms.
Particularly, we follow~\cite{realisdance, talkact} to involve 3D hand meshes as the motion signals for better hand structure recovery due to the sufficient HOI information they provided (e.g., position, hand pose, and hand size).

However, we achieved two interesting observations under the cross-object reenactment setting: \textbf{1)} Generated objects fail to preserve their original structures and textures and hand synthesis suffers from severe deformation and distortion. \textbf{2)} Objects with an obvious gap in shape or size may result in physically unreasonable interactions or grasping positions.
Therefore, we have conducted extensive explorations of motion instructions, hand-object restoration, and effective strategies for inference.

\noindent\textbf{Layout Instruction.}
A possible explanation for the first observation is that object synthesis lacks effective guidance, causing variations in both shape and texture.
In terms of the hand synthesis distortion, we speculate that the hand synthesis is strongly bonded with the positional instructions provided by hand meshes, thus the model tends to synthesize HOI scenes with similar hand positions, leading to the degradation in the final results.

To address these issues, we involve a set of layout representations composed of one bounding box for the object and two for the hands.
The object box can compensate for the position and size instructions that the reference image cannot provide.
The hand boxes are detected by a 2D key-point estimation approach~\cite{dwpose}, while the object box is derived from the object mask produced by the segmentation model~\cite{lisa}.
Particularly, the hand boxes are limited to squares with a fixed size, which constructs a pose-invariance and size-invariance instruction for hands. Such hand instructions enable positional information disentanglement from the motion signals and provide basic interactive information for HOI synthesis.

To reduce computational complexity, we utilize a lightweight network as the layout encoder, comprising 4 convolution layers initialized with Gaussian weights, with the final projection layer using zero convolution. The layout feature $\textbf{F}_l$ extracted by the layout encoder is then combined with the Gaussian noise $\epsilon_t$ and fed into the Denoising U-Net as the noisy latent.


\noindent\textbf{Hand-Object Interaction Restoration Module.}
Although the layout instructions already provide HOI information, it is not sufficient for satisfactory hand-object recovery. 
To generate accurate and high-quality hand gestures, we re-use 3D hand meshes $V^{h}$ reconstructed by HaMeR~\cite{hamer}. As mentioned above, in order to eliminate the over-reliance on hand positions, we apply simple augmentation on the positions of the hand meshes during training. As illustrated in Fig.\ref{pipeline}, the augmentation involves randomly shifting both hands in any direction from their original positions.

To capture robust and accurate hand pose information, we first utilize a ControlNet-like network to encode the hand mesh since it incorporates spatial and contextual cues into the generation process. The encoding process can be described as follows:
\begin{equation}
\mathbf{F}^{h}=\bm{C}\left(\mathbf{z}_t \mid I^{h}, t, \theta^{\mathrm{h}}\right),
\end{equation}
where $\bm{C}\left(\cdot, \theta^{\mathrm{h}}\right)$ denotes the ControlNet \cite{controlnet}, $\mathbf{z}_t$ is the noise latent diffused at timestep $t$, $\mathbf{F}^{h}$ is a set of features output by the down blocks and middle blocks of ControlNet.

It is widely recognized that generating human hands and object textures is a challenging task \cite{stablediffusion}. We contend that relying solely on aligned hand and object features is insufficient for accurately recovering details. Thus, we propose two global memory banks to restore diverse hand poses and object information, respectively. Specifically, we develop a Hand-Object Interaction Restoration module for generating human hands and fine-grained object textures. This is achieved by constructing separate learnable memory banks for hands and objects: $\mathbf{B}_h \in \mathbb{R}^{N_h \times C_h}$ and $\mathbf{B}_o \in \mathbb{R}^{N_o \times C_o}$.

Alongside these two memory banks, we also design corresponding Hand-Attention and Object-Attention layers integrated into the U-Net architecture. The Hand Memory Bank effectively restores human hand textures, while the Object Memory Bank is designed to store object textures during training. The attention mechanism for Hand Attention and Object Attention can be defined as follows:
\begin{equation}
    \mathbf{F}^{a} = \operatorname{Att}(\mathbf{F}, \textbf{B}, \textbf{B}) * M + \mathbf{F},  
\end{equation}

{\small
\begin{equation}
  \operatorname{Att}(\mathbf{F},\textbf{B},\textbf{B})  =  \operatorname{Softmax}\!\left(\!\frac{(\textbf{W}_{Q} \!\cdot \!\mathbf{F})(\textbf{W}_{K}\!\cdot\! \textbf{B})^\top\!}{\sqrt{d}}\!\right) \!\cdot\! (\textbf{W}_{V} \!\cdot\! \textbf{B}),
 \end{equation}
 }
among them, $\textbf{W}_Q$, $\textbf{W}_K$, $\textbf{W}_V$  represent learnable weights
for the cross-attention modules,  $\mathbf{B}$ is the hand or object memory bank and $M$ denotes the mask for the hand or object.
For object attention, $\mathbf{F}$ is the Denoising U-Net feature. 
As for hand attention, $\mathbf{F}=\mathbf{F}^{h}$, $\mathbf{F}^{a}$ is then added to the Denoising U-Net feature after hand attention.

\begin{figure}[h!]
	\centering
\includegraphics[width=1\linewidth]{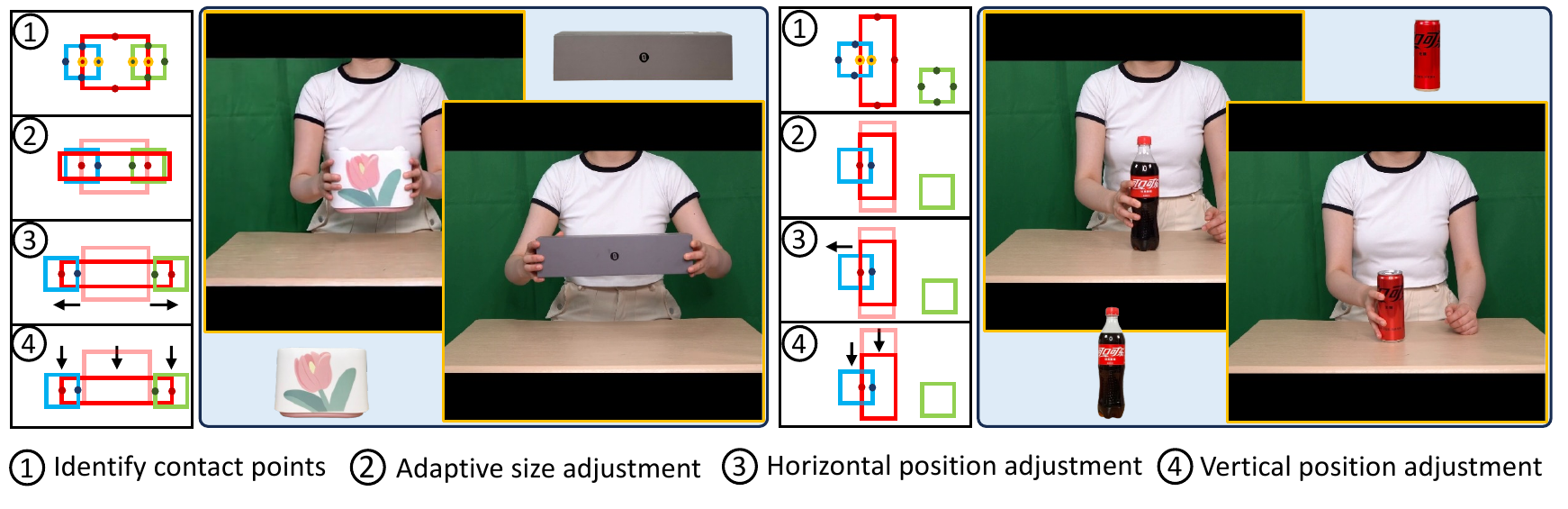} 
	\caption{Schematic diagram of adaptive layout adjustment. }
	\label{fig:adjustment}
\end{figure}

\noindent\textbf{Adaptive Layout-Adjustment Strategy.}
In terms of the second observation, we develop an adaptive strategy for layout adjustment during cross-object reenactment to produce plausible HOI physical contact relationships.
As illustrated in Fig.\ref{fig:adjustment}, the process is divided into four steps:
\textbf{1)} We initialize the centers on the four sides of each object box as potential contact points between the hand and the object. We identify the contact relationship between the hands and the box by calculating the distance from the center point of the hand box to the nearest contact point (named H2O distance). If H2O distance is less than a pre-defined threshold $\mathcal{T}$, the hand and the object are regarded as ``in contact'', otherwise they are not in contact. 
\textbf{2)} For each frame, we fix the center point of the object box and then adjust its height and width to match the size of the target object by calculating the adaptive ratio by frame. 
\textbf{3)} Then we horizontally adjust the position of each hand box to maintain the original H2O distance.
\textbf{4)} Finally, we keep moving the object box until its bottom is consistent with the original box bottom. 
Following these four steps, we can effectively avoid floating objects and generate physically reasonable hand-object interactions, especially when handling objects with obvious gaps in size and shape.

\begin{table*}[t]
\setlength{\tabcolsep}{6pt}
\belowrulesep=0pt\aboverulesep=0pt
 \centering
  
 \renewcommand{\arraystretch}{1.1}
    \begin{tabular}{c|c|ccc|cccccc}
        \hline
        \multirow{2}*{Dataset}& \multirow{2}*{Method} & \multicolumn{3}{c|}{Cross-Reenactment} & \multicolumn{6}{c}{Self-Reenactment} \\ 
          \cline{3-11} 
           & & \makecell[c]{hand \\fid.}$\uparrow$ & \makecell[c]{subj.\\cons.}$\uparrow$ &\makecell[c]{mot.\\smth.}$\uparrow$ &PSNR $\uparrow$& FID $\downarrow$ &\makecell[c]{hand\\agr.} $\uparrow$ &\makecell[c]{hand \\fid.} $\uparrow$ & \makecell[c]{subj.\\cons.} $\uparrow$ &\makecell[c]{mot.\\smth.} $\uparrow$\\
         \hline
               \multirow{4}{*}{HOI4D}&AnyV2V  
               &0.183 &0.591 &0.952 &28.821
               &185.961 &0.206 &0.291 &0.877 &0.982 \\  
               \multirow{4}{*}{}&VideoSwap 
               &0.924 &0.907 &\textbf{0.992} &31.964
               &158.964 &0.611 &0.987 &0.915&0.990  \\  
               \multirow{4}{*}{}&HOI-Swap 
               &0.994 &0.911 &0.990 &31.528
               &30.152 &0.754 &0.993 &0.902&0.988  \\ 
               \cline{2-11}
               \multirow{4}{*}{}
               &Re-HOLD \cellcolor{gray!40} 
               &\textbf{0.994} \cellcolor{gray!40} 
               &\textbf{0.915} \cellcolor{gray!40} 
               &0.991 \cellcolor{gray!40} 
               &\textbf{31.984} \cellcolor{gray!40} 
               &\textbf{26.583} \cellcolor{gray!40} 
               &\textbf{0.826}  \cellcolor{gray!40} 
               &\textbf{0.993} \cellcolor{gray!40} 
               &\textbf{0.916} \cellcolor{gray!40} 
               &\textbf{0.991} \cellcolor{gray!40} \\  
            \hline
            \multirow{6}{*}{Ours}& AnyV2V  
               &0.934 &0.829 &0.983 &30.166
               &116.084 &0.223 &0.981 &0.931&0.992  \\
               \multirow{6}{*}{}&VideoSwap 
               &0.936 &0.922 &0.992 &32.903
               &100.840 &0.625 &0.983 &0.943 &0.993  \\    
               \multirow{6}{*}{}&AnimateAnyone 
               &0.983 &0.950 &0.991 &32.611
               &26.361 &0.698 &0.990 &0.951 &0.992\\  
               \multirow{6}{*}{}&RealisDance 
               &0.989 &0.948 &0.991 &32.784
               &26.337 &0.749 &0.992 &0.951 &0.993 \\  
               \multirow{6}{*}{}&HOI-Swap 
               &0.994 &0.944 &0.994 &31.634
               &30.932 &0.725 &0.992 &0.949&0.994 \\ 
            \multirow{6}{*}{}
                &Re-HOLD   \cellcolor{gray!40} 
                &\textbf{0.994} \cellcolor{gray!40} 
                &\textbf{0.955} \cellcolor{gray!40} 
                &\textbf{0.994} \cellcolor{gray!40} 
                &\textbf{33.451} \cellcolor{gray!40} 
                &\textbf{19.021} \cellcolor{gray!40} 
                &\textbf{0.773} \cellcolor{gray!40} 
                &\textbf{0.993} \cellcolor{gray!40} 
                &\textbf{0.953} \cellcolor{gray!40} 
                &\textbf{0.995}  \cellcolor{gray!40}\\  
        \hline
         \end{tabular}
        \caption{ Quantitative results of our approach compared with SOTAs. `Cross-Reenactment' means the target object is different from the original object while `Self-Reenactment' is the self-reconstruction result on the test set.}
\label{tab:exp}
\end{table*}

\subsection{Training Objectives}
\label{train}

Our framework is developed from Stable Diffusion (SD) \cite{stablediffusion}, which employs an autoencoder to diffuse and denoise within the latent space.
In the training phase, the diffusion process involves encoding the image to latent space $\mathbf{z}_0=\mathcal{E}(\mathrm{x})$, and the random Gaussian noise is gradually added to $\mathbf{z}_0$ with $T$ diffusion steps. 
Typically, the training object is to predict the added noise at various time steps through a learnable denoising U-Net, formulated as follows:
\begin{equation}
\mathbf{L}=\mathbb{E}_{\mathbf{z}_t, c, \epsilon, t}\left(\left\|\epsilon-\epsilon_\theta\left(\mathbf{z}_t, c, t\right)\right\|_2^2\right),
\end{equation}
where $\epsilon_\theta$ denotes the denoising U-Net,
$c$ is the conditional embeddings. 
During inference, $\mathbf{z}_T$ is sampled from random Gaussian distribution with the initial timestep $T$ and is denoised back into $\mathbf{z}_0$. 
Finally, the decoder $\mathcal{D}$ reconstructs $\mathbf{z}_0$ to yield the generated image.

In our work, we adopt a two-stage training strategy to perform image-level HOI modeling and temporal HOI consistency modeling separately.
The first stage is image-level HOI modeling, which focuses on accurately establishing HOI based on given image-level conditions.
In this stage, we keep the VAE encoder fixed while training the remaining network components, excluding the temporal attention mechanism. 
Notably, towards the end of the first training phase, we place a special emphasis on hands and objects by only calculating the L1 loss of the corresponding region as the final loss every 10 iterations.
In the second stage, we incorporate the temporal layer into the previously trained network to model the temporal consistency of the generated video frames.
During this stage,  the input is consecutive video frames and we only train the temporal layer while fixing the weights of the rest of the network.

\begin{figure*}[t!]
	\centering
\includegraphics[width=1\textwidth]{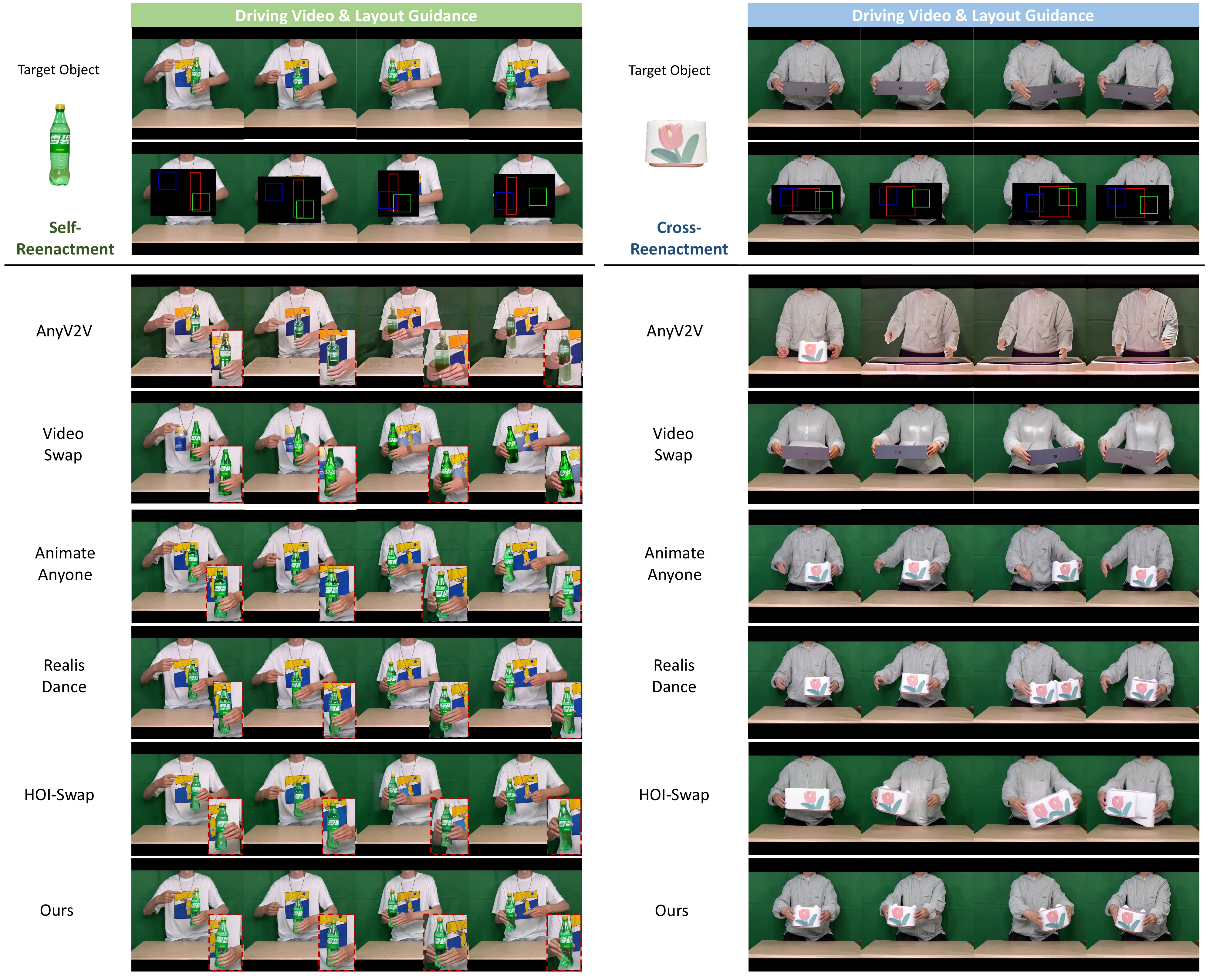} 
	\caption{\textbf{Qualitative results compared with other methods.} Our approach achieves high-fidelity HOI details and satisfactory image quality in both self-reenactment and cross-object reenactment settings.}
	\label{exp}
\end{figure*}

\section{Experiments and Results}
\label{experiment}
\subsection{Experimental Settings}
\label{experiment set}
\textbf{Dataset.}
For our training, we collect a dataset consisting of 9 subjects with 14 objects. 
All videos are segmented into clips of 5 seconds each. 
For each subject data, we randomly select two objects that the subject has not seen as the test set. 
To enrich the diversity of the dataset, we follow HOI-Swap \cite{hoi-swap} to leverage a large-scale egocentric dataset HOI4D \cite{hoi4d} for training. 
To meet our task setting, we only select videos that include a single object in HOI4D. 

\noindent\textbf{Implementation Details.} 
For data pre-processing, we first crop out the hand-object interaction region excluding the human face based on the key points of the DWPose \cite{dwpose}. 
We then input cropped frames into HaMeR \cite{hamer} to predict the MANO \cite{mano} model parameters for rendering the 3D hand mesh. 
The hand box is also extracted by DWPose \cite{dwpose} for the layout construction. We utilize LISA \cite{lisa}, a language-guided segmentation model for the extraction of the object mask. All video clips are pre-processed at a frame rate of 25 FPS with 512×512 resolution. The hand and object feature bank size is empirically set to 512. Since the size of the hand pose varies, while the hand box size remains the same, so we set $\mathcal{T}$ to half the size of the hand box in addition to 20. Following
AnimateAnyone \cite{animateanyone}, we initialize the Denoising Branch and Reference Branch using Stable Diffusion V1.5 parameters. During inference, we use a DDIM sampler for 30 denoising steps. All experiments were completed on 4 A800s with a learning rate of 1e-5. For the first training stage, the batch size is 48 and F is 1. The training step is 100k, which takes about three days. For the second stage, batch size and F are set to 1 and 24 respectively with 50k training steps, taking about 2 days. 


\noindent\textbf{Evaluation Metrics.} To demonstrate the effectiveness of our method, we comprehensively measure the self-reenactment results as well as the cross-reenactment results. Following HOI-Swap \cite{hoi-swap}, we employ the HOI hand agreement to measure spatial alignment in the hand region, hand fidelity, subject consistency, and motion smoothness to evaluate general video quality. We exclude the HOI contact agreement metric due to the incorrect detection of objects. We also measure the quality of generated images from pixel space and feature space using PSNR and FID \cite{FID}. Hand agreement score, PSNR, and FID are only used for evaluating self-reenactment results due to the lack of ground truth in the cross-reenactment setting.

\subsection{Comparison with Other Methods}
\textbf{Quantitative Results.} For a more thorough comparison, we conduct two experimental settings, including self-reenactment and cross-object reenactment. Self-reenactment is performed only on the test set, which consists of unseen objects for each person. For cross-object reenactment, we displace the objects of the training set to the ones in the test set. The quantitative results of our methods compared with SOTAs are shown in Table \ref{tab:exp}. AnyV2V \cite{anyv2v} and VideoSwap \cite{videoswap} are the state-of-the-art video editing methods. AnimateAnyone \cite{animateanyone} and RealisDance\cite{realisdance} focus on generating human motions without interacting with objects. HOI-Swap \cite{hoi-swap} aims to swap the objects with HOI awareness. As illustrated in Table \ref{tab:exp}, 
Our method achieves top performance in both self-reenactment and cross-object reenactment across most metrics, proving its effectiveness. For instance, our method significantly outperforms other methods in PSNR and FID metrics, demonstrating the superiority of our approach in image generation. In addition, we obtain the best hand fidelity and hand agreement metrics indicating that Re-HOLD can synthesize accurate hand poses. Additionally, our approach maintains the best subject consistency, ensuring high-fidelity object textures in generated videos.
Benefiting from the adaptive layout strategy, we can generate appropriate HOI details during inference. 

\noindent\textbf{Qualitative Results.} For qualitative comparison, we provide results under both self-reenactment and cross-object reenactment settings. As shown in the left part of Fig.\ref{exp}, our method can generate realistic object texture though it is unseen by the person while other approaches fail to do so. Note that cross-object reenactment is more challenging due to the varied object shapes and sizes, but we still achieve proper HOI details and satisfactory image quality as indicated in the right of Fig.\ref{exp}. The results of the other method either appear as two objects in the image or generate another one instead of the target. In conclusion, our method fully exploits human-object interaction (HOI) information through adaptive layout guidance and the HOI Restoration Module, thereby enhancing video quality to meet the demands of various cross-reenactment scenarios.

\noindent\textbf{Human Evaluation.} We conduct a user preference study on our collected dataset to evaluate the performance of human-centric HOI video generation. There are 20 samples and 15 human voters in total. For each sample, we randomly present six video results generated with Re-HOLD and other SOTA methods to the human voter. The human voters are required to estimate the video results in three aspects: 
a) HOI Consistency: Does the video accurately reproduce the Human-Object Interaction in the driving video? 
b) Object Appearance Consistency: Does the object in the video have a consistent appearance with the target object? 
c) Temporal Consistency: How is the temporal coherence of this video? The rating score ranges from 1 to 5 and higher scores indicate better preference. The displayed result represents the ratio of the obtained score to the overall score. As shown in Table \ref{tab:userstudy}, our method achieves the highest scores compared with its counterparts.

\begin{table}[t!]
\setlength{\tabcolsep}{3pt}
\begin{minipage}{\columnwidth}
\begin{center}
\begin{tabular}{cccc}
  \toprule
  Method & \makecell[c]{HOI \\Consistency} & \makecell[c]{Object \\Consistency} & \makecell[c]{Temporal \\Consistency} \\
  \midrule
      AnyV2V       & 0.38 & 0.32 & 0.38 \\
VideoSwap      & 0.52 & 0.22 & 0.44  \\
    AnimateAnyone   & 0.72 & 0.58 & 0.42\\
    RealisDance    & 0.68 & 0.74 & 0.28 \\
    HOI-Swap    & 0.76 & 0.40 & 0.44 \\
  \textbf{Re-HOLD}  & \textbf{0.92}  & \textbf{0.92} &\textbf{0.88} \\
  \bottomrule
\end{tabular}
\caption{User study of Re-HOLD and other SOTA methods. }
\label{tab:userstudy}
\end{center}
\end{minipage}
\end{table}
\begin{figure}[h!]
	\centering
\includegraphics[width=0.9\linewidth]{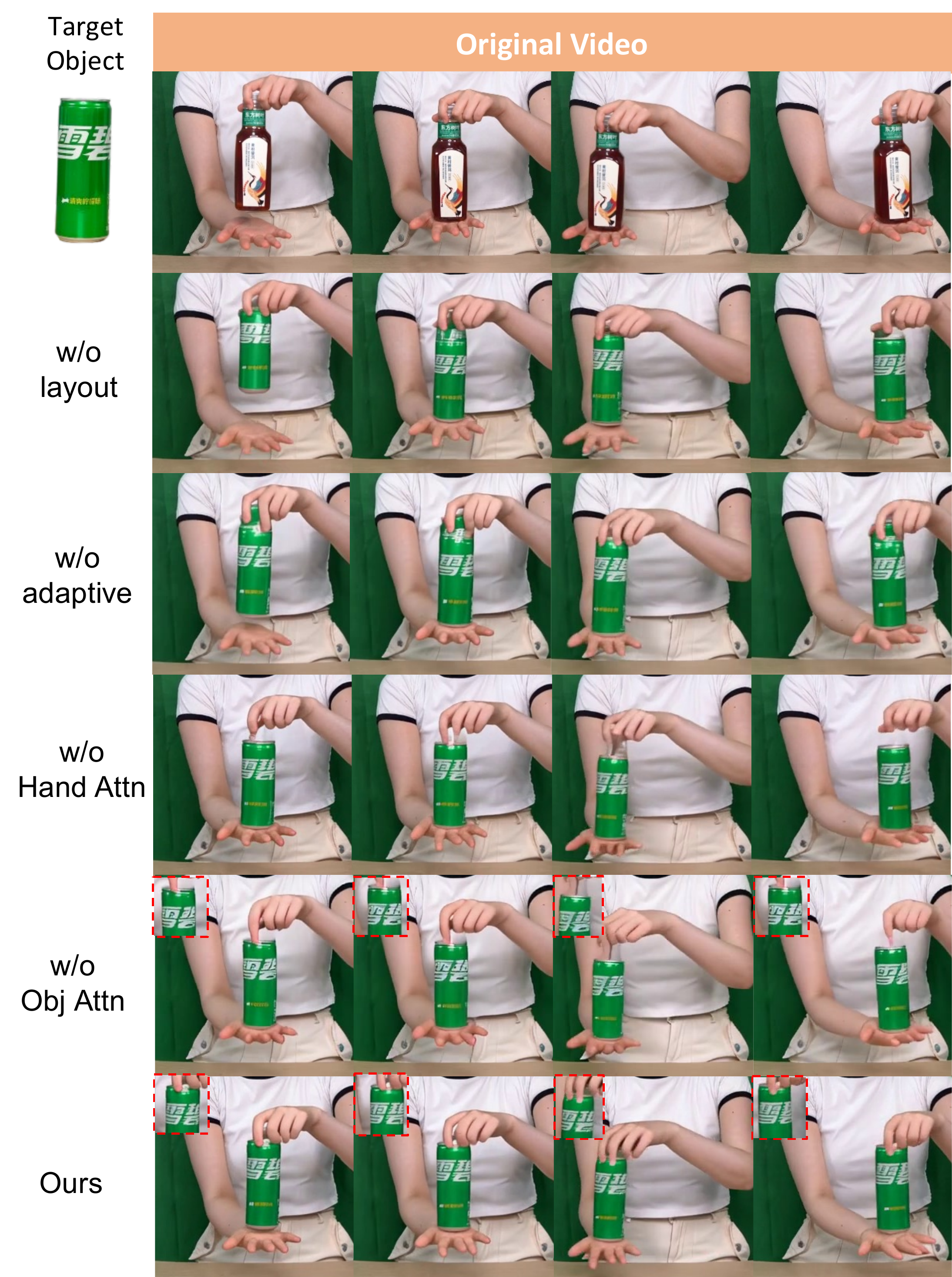} 
	\caption{Qualitative ablation results of cross-reenactment when removing different components in our framework.}
	\label{fig:ablation}
\end{figure}

\subsection{Ablation Study}
To better demonstrate the effectiveness of different components of our framework, we conduct experiments for all the variants in our ablation studies, which are defined below. 
1) ``w/o layout": We directly use the 3D hand mesh renderings as the driving signal instead of the proposed layout guidance and the ControlNet is removed accordingly. 
%
%
2) ``w/o adaptive strategy": The adaptive layout adjustment strategy is not applied under the cross-reenactment setting. 
3) ``w/o hand attention": We remove the hand attention layers of the HOI Restoration Module from the Denoising U-Net. 
4) ``w/o obj attention": We remove the object attention layers of the HOI Restoration Module from the Denoising U-Net. 

For quantitative results, we report the hand fidelity, and subject consistency metrics in Table \ref{tab:ablation}. It is observed that hand attention can enhance the hand gesture quality, and object attention brings more detailed information about objects. Our result in the subject consistency metric is higher than ``w/o adaptive strategy" indicating the significance of this strategy. Note that the adaptive layout adjustment strategy is only performed in the cross-reenactment setting. 
Additionally, we present qualitative comparisons of cross-object reenactment to verify the effectiveness of the proposed module. As shown in Fig.\ref{fig:ablation}, the object is deformed without the layout guidance, demonstrating the validity of our proposed method.

\begin{table}[!t]
\setlength{\tabcolsep}{1pt}
\begin{minipage}{\columnwidth}
\begin{center}

\begin{tabular}{lcccc}
  \toprule
  \multirow{2}*{Variations} & \multicolumn{2}{c}{Self-Reenactment} & \multicolumn{2}{c}{Cross-Reenactment} \\ 
          \cmidrule(lr){2-3}\cmidrule(lr){4-5}
          & \makecell[c]{hand \\agr.}$\uparrow$ & \makecell[c]{subj.\\cons.}$\uparrow$ & \makecell[c]{hand \\fid.}$\uparrow$ & \makecell[c]{subj.\\cons.}$\uparrow$  \\
  \midrule
  w/o layout            & 0.753 & 0.950 & 0.993 &0.950\\
  w/o adaptive strategy     & - & - & 0.992 & 0.952 \\
    w/o hand attention   & 0.756 & 0.952 & 0.992 & 0.953\\
    w/o obj attention    & 0.767 & 0.951 & 0.994 & 0.952\\
  \textbf{Ours}  & \textbf{0.773}  & \textbf{0.953} &\textbf{0.994} &\textbf{0.955}\\
  \bottomrule
\end{tabular}
\caption{Ablation study.}
\label{tab:ablation}
\end{center}
\end{minipage}
\vspace{1mm} 
\end{table}

\section{Discussion and Conclusion}
\noindent\textbf{Conclusion.}
In this paper, we propose the video reenactment framework Re-HOLD, which achieves realistic and reasonable  Human-Object Interaction (HOI) via an adaptive Layout-instructed Diffusion model.
We first specialize in layout representations of hands and objects for effective hand-object disentanglement.
Accordingly, we introduce a Hand-Object Interaction Restoration module to perform structure reshaping and texture refinement via two memory banks.
To further bridge the gap between diverse objects under the cross-reenactment setting, we implement an adaptive layout adjustment strategy that enables the generation of plausible hand-object physical contacts.
Both quantitative and qualitative assessments have demonstrated our framework's superiority over existing methods.

\noindent\textbf{Limitations.}
Despite the success of our framework, we also recognize some limitations during the exploration. Our dataset is specifically designed to capture fundamental hand movements used for object display in live-streaming scenarios. As a result, our framework produces less satisfactory results when handling 3D object manipulation cases, such as generating a multi-view video of an object. This will be a focus of our future work.

\noindent\textbf{Broader Impact.} 
However, the potential for misuse of this technology is a significant concern. 
We will take measures to strictly monitor the content our model generates, restricting access to research-oriented applications only. 
We believe the responsible use of our model can foster positive societal development in academic research and everyday life.

\section*{Acknowledgment}
 This work was partially supported by the Yunnan provincial major science and technology special plan projects under Grant 202403AA080002 and the National Natural Science Foundation of China under grant 62372341.
{
    \small
    \bibliographystyle{ieeenat_fullname}
    \bibliography{main}
}


\end{document}